\documentclass{article}
\usepackage[T1]{fontenc}
\usepackage{spconf,amsmath,graphicx}
\usepackage[hidelinks,pdfversion=1.7]{hyperref}
\usepackage{booktabs}
\usepackage[protrusion=false]{microtype}
\usepackage{balance}
\hypersetup{
  pdftitle={AUDITING ACTION SETTLEMENT IN LLM AGENT ENVIRONMENTS: ORDER, PROGRESS, AND REPLAY},
  pdfauthor={Haotian Chen, Bowen Ye, Yuning Zhang, Jingkun Yu},
  pdfkeywords={LLM agents, multi-agent simulation, action settlement, order sensitivity, replay}
}

\title{AUDITING ACTION SETTLEMENT IN LLM AGENT ENVIRONMENTS: ORDER, PROGRESS, AND REPLAY}

\name{Haotian Chen$^{1,\dagger}$ \qquad Bowen Ye$^{2,\dagger,\ast}$ \qquad Yuning Zhang$^{1}$ \qquad Jingkun Yu$^{3}$}
\address{$^{1}$ School of Cyber Science and Technology, University of Science and Technology of China \\
$^{2}$ School of Education, Shanghai Jiao Tong University \\
$^{3}$ SWJTU-Leeds Joint School, Southwest Jiaotong University \\
$^{\dagger}$ Equal contribution \quad $^{\ast}$ Corresponding author: \texttt{9603213113@sjtu.edu.cn}}

\begin{document}
\maketitle

\begin{abstract}
Concurrent actions in large language model (LLM) agent environments require
arbitration even when each proposal is individually valid. We implement a typed
snapshot--settlement contract and audit three distinct properties: order
sensitivity, useful progress, and replay consistency. Five settlement policies
are tested in 28,800 exhaustive permutation trials and 2,160 scripted multistep
episodes. Joint policies are spatially order-invariant conditional on fixed
priorities, yet conservative rejection completes only 31.25\% of agents in a
six-agent doorway task versus 90.28\% for random tickets; the paired improvement
is 59.03 percentage points (95\% bootstrap interval: 50.00--68.06). All policies preserve
the tested spatial constraints, and priority arbitration still misses the
independent small-instance optimum. A separate full-state journal audit exactly
replays 156 checkpoints and rejects 1,332 constructed corruptions with a
retained terminal anchor. The evidence concerns execution semantics, not
human realism or long-run fairness.
\end{abstract}


\section{Introduction}
\label{sec:intro}

LLM agents support interactive environments \cite{park2023generative}, opinion
dynamics, and social learning
\cite{yazici2026opinion,firsova2026nonbayesian,chuang2024opinion}.
Measurements depend on arbitration as well as agent policies.
Individually valid requests for one cell or capacity-limited object may be
jointly infeasible. Response-order commitment can change access and progress
with fixed intentions; rejecting every conflict removes order effects but
may prevent useful activity.

The influence--reaction model already separates attempts from a joint
environment reaction \cite{ferber1996influences}. PettingZoo discusses
order-biased same-cell conflicts \cite{terry2021pettingzoo}; Concordia grounds
attempts through game-master events \cite{vezhnevets2023concordia}, while
executable language environments constrain transitions
\cite{shridhar2021alfworld,seo2026simuhome}. A simultaneous API alone does not
specify arbitration. We claim neither synchronization nor intent/event
separation as a new architecture.

We contribute (i) a typed contract and scoped spatial invariance argument;
(ii) five matched arbitration controls, an independent feasible-subset oracle,
and held-out multistep stress conditions; and (iii) serialized-state replay
with fault injections and explicit trust limits. We study spatial execution,
not social validity. Multi-agent pathfinding (MAPF) likewise requires explicit
conflict and goal-occupancy rules \cite{stern2019mapf}; we settle one-step
proposals, not joint paths. Generic LLM planning has documented limitations
\cite{valmeekam2023planning}; MAPF-GPT instead learns from expert paths
\cite{andreychuk2025mapfgpt}, unlike our prompted agents.

\begin{figure*}[t]
\centering
\includegraphics[width=\textwidth]{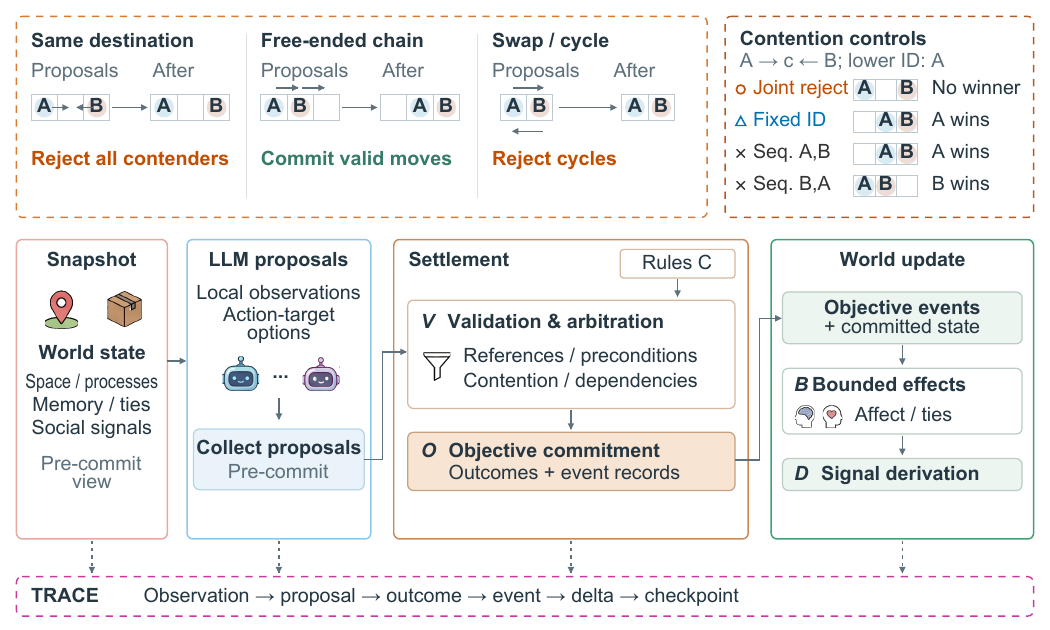}
\caption{Snapshot--settlement contract. Upper panels illustrate joint rejection,
fixed ID, and sequential controls; A has lower ID and A,B/B,A gives sequential
order. Below, snapshot-based proposals undergo validation $V$, commitment $O$,
bounded subjective updates $B$, and signal derivation $D$. Dashed arrows denote
logging, not execution. Journal coverage starts before pending-action/process
advancement. Invariance covers positions only.}
\label{fig:tick}
\end{figure*}

\section{Settlement and Replay Contract}
\label{sec:method}

\subsection{Typed proposals and objective commitment}

In Fig.~\ref{fig:tick}, after pending actions and processes advance, a frozen state $\hat{x}_t$
supplies all scheduled observations. State includes locations, occupancy,
processes, queues, events, memories, relationships, affect, and derived signals.
Each agent selects an explicitly bound action--target option and returns
$p_{i,t}=(u_{i,t},\tau_{i,t},d_{i,t},s_{i,t})$: an action identifier, target,
duration, and short recorded summary. Role, visibility, activation, and process
rules determine candidates; model text cannot register new actions. With the
proposal collection $P_t$, world constraints $C$, and objective events $\mathcal E_t$,
\begin{align}
(\mathcal E_t,x_t^{O})&=O(\hat{x}_t,V(P_t;\hat{x}_t,C)),\\
x_{t+1}&=D(B(\hat{x}_t,x_t^{O},\mathcal E_t)).
\label{eq:transition}
\end{align}
Validation $V$ checks references, preconditions, spatial and capacity rules; $O$
commits outcomes. Subjective update $B$ may use model estimates but bounds
committed affect and relationship values; $D$ derives signals. Subjective
estimates do not overwrite objective positions. Long actions prevent
redecision until completion. Invalid decision responses receive one repair,
then a recorded idle fallback.

Movement uses adjacent cardinal steps. Occupied destinations require their
occupants to move along a dependency chain terminating at a free cell. Swaps
and directed cycles are forbidden by our chosen semantics, not a universal
MAPF safety rule. The historical candidate selector hides occupied targets;
an explicit opt-in exposes them for simultaneous-chain proposals. Joint
policies group object requests before mutation: accepted new occupants cannot
exceed capacity minus snapshot occupancy. They do not credit same-tick releases;
sequential commitment may credit an earlier release.

\subsection{Five controlled policies}

All policies retain validation. \emph{Joint rejection} rejects every contender
for an oversubscribed destination/resource. \emph{Sequential} commits proposals
one by one against updated state. Joint-priority controls (\emph{fixed ID},
\emph{random tickets}, and \emph{rotating slots}) select contenders before the
same dependency checks. Tickets are unique ranks drawn before inspecting arrival
order and stored in the trace. Rotation advances a stable actor-slot ranking
each tick. Rejected actors remain occupants; greedy winners are not reconsidered
if downstream dependencies fail. Rejection remains the runtime default.

Let $R_\rho$ be settlement and $\Pi_L$ the position projection. Assume unique
initial occupancy, one move/idle proposal per agent, deterministic
snapshot validation, no other position updates, and fixed priorities where
applicable. For every joint policy and input permutation $\pi$,
\begin{equation}
\Pi_L R_\rho(\hat{x},P)=\Pi_L R_\rho(\hat{x},\pi(P)).
\label{eq:invariance}
\end{equation}
\emph{Argument.} Destination groups and priority winners do not depend on list
order. After arbitration, a move succeeds precisely when its
occupancy dependency reaches a free cell without a rejected mover or cycle.
This predicate defines an order-independent set of moves with distinct final
destinations. It does not imply full-state invariance or maximal throughput.
Fixed IDs can privilege an identity. Random tickets are exchangeable across
draws, not identical under independently redrawn relabelings. Coupled
relabeling must carry ranks/slots with physical actors. None of these
observations proves long-run fairness.

\subsection{Complete-state journal}

A journal records the initial runtime state, tick-begin \emph{before} pending
actions/processes advance, and commit deltas. Independent replay checks
sequence numbers, pre/post state hashes, and a record hash chain, then compares
reconstructed state with checkpoints. It applies recorded subjective outcomes
without resolver/LLM calls. Continuation requires a terminal-state match;
legacy resumes declare their coverage start, not certification of missing history.

A separately retained terminal detects missing complete suffixes. A colocated
editable terminal is not authenticated: an adversary can rewrite both.
Replay tests serialization consistency, not scientific correctness. Coverage
excludes static world/configuration, separate text logs, and model/RNG state.

\section{Evaluation}
\label{sec:eval}

\subsection{Frozen design and independent checks}

We separate scripted mechanism tests, production-engine integration, and
archived LLM execution. New protocols and source hashes are frozen before
formal runs; smoke seeds are disjoint. The spatial benchmark uses 24 formal
seeds, three task families, three loads, standard and held-out reflected
layouts, and five policies: 2,160 episodes. Reflection is a geometric variant,
not a new task family. Free-chain and doorway loads are 2/4/6; hub-service
loads are 2/3/4. Hub agents visit a shared center then return home. All agents
stay at final goals; the horizon is 24 ticks. A common scripted policy follows
shortest static paths with seed-based tie-breaking, replanning from each
method's own state after divergence. Input orders are matched across policies.
This policy is neither an LLM nor a coordination planner.

Completion is the within-episode fraction of agents occupying their final goal
at the horizon. Restricted mean completion time averages per-agent arrival
ticks, assigning unfinished agents the horizon; it is not makespan. Paired
differences use 24 seed episodes per task, load, and layout stratum and 2,000
bootstrap resamples. Intervals are pointwise, not multiplicity-corrected;
ticks, agents, permutations, and reflected copies are not independent
replicates.

An independent feasibility checker validates every spatial benchmark
transition against distinct occupancy, cardinal-step, and cycle constraints,
without runtime validation. Exhaustive subset enumeration supplies the maximum
feasible move count for fixed state/proposals, including 120 random instances
under every policy. This optimum is not a globally optimal trajectory.

\begin{table}[t]
\centering
\caption{Standard-layout high-load results, 24 seeds per cell.
Door/hub: mean agent completion (\%). Chain: restricted mean per-agent
completion time (ticks). All policies satisfy tested spatial constraints.}
\label{tab:progress}
\small
\setlength{\tabcolsep}{2.5pt}
\begin{tabular}{lrrr}
\toprule
Policy & Door ($n=6$) & Chain ($n=6$) & Hub ($n=4$)\\
\midrule
Joint rejection & 31.25 & 3.167 & 0\\
Sequential & 89.58 & 5.208 & 100\\
Fixed ID & 91.67 & 3.167 & 100\\
Random tickets & 90.28 & 3.167 & 100\\
Rotating slots & 88.89 & 3.167 & 100\\
\bottomrule
\end{tabular}
\end{table}

\begin{figure}[!t]
\centering
\includegraphics[width=\columnwidth]{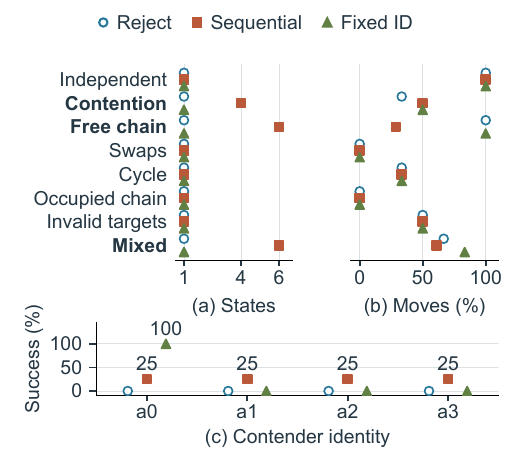}
\caption{Three-policy one-tick diagnostic: eight six-agent motifs $\times$ 720
orders (17,280 scripted trials). (a) Distinct final position states;
(b) agents moving in one tick (\%), not multistep goal completion;
(c) contender success over all orders, not long-run fairness.
All policies have zero tested spatial violations and exact position-delta replay.}
\label{fig:order-audit}
\end{figure}

\subsection{Order sensitivity and useful progress}

Eight six-agent motifs cover independent moves, contention, free chains,
swaps, cycles, stationary-terminated chains, invalid targets, and mixed cases.
Enumerating all 720 orders under five policies gives 28,800 trials. Every
joint policy yields one position state per motif with fixed tickets/slots;
sequential commitment yields four in contention and six in the free chain.
Figure~\ref{fig:order-audit} visualizes the three-policy one-tick diagnostic
(rejection, sequential, and fixed ID); the 28,800-trial audit additionally
includes saved tickets and rotating slots.
All methods preserve the tested spatial constraints. Order sensitivity need
not imply corruption, so these tests establish no safety advantage over the
validated sequential control.

The multistep benchmark has 21,213 ticks and 83,400 agent decisions with zero
tested spatial violations. Useful progress nevertheless differs substantially
(Table~\ref{tab:progress}). For the standard six-agent
doorway, random tickets improve completion over rejection by 59.03 percentage
points (paired 95\% bootstrap interval [50.00, 68.06]); the reflected difference
is 63.89 points [55.56, 72.22]. Fixed-ID and rotating priorities are also strong controls:
random tickets do not dominate them. All policies finish free chains, but
joint commitment reduces completion time. In four-agent hub service,
rejection completes no agents within the horizon whereas all other policies
complete every agent. Eliminating list-order effects is therefore insufficient
for useful execution.

Nor is greedy priority optimal. Across 120 random instances, the oracle admits
495 moves in total; rejection, sequential, fixed-ID, tickets, and rotation
accept 465, 451, 487, 488, and 487. Both fixed-ID and tickets miss the maximum
in six instances.

\subsection{Waiting and allocation}

A prespecified starvation proxy counts agents with at least eight consecutive
failed move attempts, including agents that later finish. In the same standard
six-agent doorway, the fractions are 68.75\%, 4.17\%, 7.64\%, 6.94\%, and
8.33\% for rejection, sequential, fixed-ID, tickets, and rotation. Sequential
has the lowest proxy here; ticket exchangeability does not eliminate extended
waiting. All-agent episode success is respectively 0\%, 50\%, 62.5\%, 62.5\%,
and 50\%, lower than mean agent completion in Table~\ref{tab:progress}.
Separately, 1,000 four-contender ticket draws give wins of 254/273/224/249.
Neither that allocation diagnostic nor a finite waiting threshold proves
long-run fairness.

\subsection{Resource and full-state replay audits}

Snapshot-valid resource actions can jointly overbook capacity. A regression
audit submits four users to capacities 1--4 over every order. The preserved
legacy resolver overbooks all 72 oversubscribed cases; grouped rejection and
validated sequential commitment each overbook none of 96 cases. Rejection
admits zero users under contention whereas sequential admits the capacity.
This repairs a runtime defect; it does not establish a novel allocation rule.

Twelve scripted production-engine runs exercise movement, object occupancy,
running-action completion, queues, processes, affect, relationships, and
memories. Across 144 committed ticks, independent replay matches all 156
full-state checkpoints and 12 final states without resolver/model calls.
With an original retained terminal, it rejects 1,332/1,332 constructed record
deletions, duplicates, adjacent swaps, modifications, and suffix truncations.
Without that terminal, all 144 complete prefixes pass. Rehashing modified
metadata also passes internal checks in 12/12 cases but fails against original
anchors. These negative controls explicitly rule out an authentication claim.

\subsection{Production integration and LLM evidence}

Sixty scripted production-engine controls cross three seeds, two tasks, two
candidate modes, and five policies; all 632 checkpoints replay. Exposing
occupied targets reduces free-chain joint-policy completion from six to four
ticks; sequential remains at six. Doorway rejection completes one of three
agents in either mode; other policies finish all three. With occupied targets
exposed, tickets finish in 11/12/11 ticks over the three seeds, versus
rejection's 1/3 at horizon 20. These are scripted controls, not LLM predictions.

Thirty archived LLM runs cover six fixed worlds (6--23 agents), five runs each;
all finish, with world-mean fallback 0.035\%. Observation/default-position
reconstruction admits 971/1,080 ticks and excludes 109 observation mismatches.
The admitted ticks contain 5,379 LLM proposals, three fallbacks, 150 moves, and
\emph{no valid destination contention}. All 36,135 five-policy one-step
interventions give identical positions, establishing neither closed-loop
arbitration benefits nor historical full-state replay. Artifacts and exclusions
remain local, not a public release.

Live development uses gateway aliases Gemini-2.5-Flash, Qwen2.5-32B, and Luna
(upstream identities unverified). An ambiguous region/coordinate interface
leaves 1/3 agents at goal after six ticks (10 calls, seven replayed checkpoints).
With explicit own-coordinate goal status, free chains under rejection finish
3/3 in 4/6/7 ticks for Gemini/Luna/Qwen (49 calls, zero fallbacks, 20 replayed
checkpoints). Reflected Qwen/rejection and Luna/tickets pilots each finish 3/3
in eight ticks (24/20 calls, nine checkpoints each, zero fallbacks). Each entry
is one episode; all five corrected episodes lack contention. Different
versions, horizons, and unseeded sampling preclude causal interface/model comparisons.

Five interrupted smoke/pilot batches complete 0/3, 1/12, 0/12, 1/4, and 0/3
scheduled episodes (2/69/14/37/21 API attempts, including failures).
The final Luna/ticket bottleneck commits six ticks: 17 accepted moves, one
blocked move, and one two-agent contested tick. It stops on HTTP 404 after 18
successful calls and two pre-submission timeouts. Its 0/3 arrivals are not a
20-tick failure: shortest paths require at least eight ticks before any arrival.
Missing outcomes are neither imputed nor pooled across restarts. A balanced
formal LLM policy comparison remains absent.

\section{Conclusion and Limitations}

Order invariance, compliance, progress, and replay are distinct. Priority
improves progress in tested scripted bottlenecks, without guaranteeing
optimality or fairness. Replay verifies consistency, not authenticity.
Small spatial tasks and unconfirmed transfer to balanced LLM comparisons
limit generalization. We claim neither human validity, broad framework
superiority, nor full-state order invariance.

\clearpage
\section{Compliance with Ethical Standards}

The study uses synthetic agents and no new human-subject experiment. LLM
simulation can reproduce stereotypes or be mistaken for evidence about real
groups. We preserve failures and exclusions and make no human-realism or
high-stakes decision claim; application to people requires external validation.

\begingroup
\balance
\interlinepenalty=10000
\bibliographystyle{IEEEbib}
\bibliography{strings,references}
\endgroup
\end{document}